\documentclass[runningheads]{llncs}

\usepackage{eccv}

\usepackage{eccvabbrv}

\usepackage{graphicx}
\usepackage{booktabs}

\usepackage[accsupp]{axessibility}  

\usepackage{hyperref}

\usepackage{orcidlink}

\begin{document}

\title{DualFormer: An Efficient MetaFormer with Parallel Token Mixers and Polynomial Kernel Mixing} 

\titlerunning{DualFormer}

\author{
Yuto Yokoi\orcidlink{0009-0005-7404-8749}\inst{1}\and
Kazuhiro Hotta\orcidlink{0000-0002-5675-8713}\inst{1}
}

\authorrunning{Y. Yokoi and K. Hotta}

\institute{
Meijo University,Japan\\
\email{253427028@ccalumni.meijo-u.ac.jp}, \email{kazuhotta@meijo-u.ac.jp}
}

\maketitle

\begin{abstract}
Image classification models benefit from capturing local features and global context. While MetaFormer based architectures have achieved strong performance, most rely on a single Token Mixer, which may limit representation diversity and the ability to model complementary feature interactions efficiently.
In this paper, we introduce a Polynomial Kernel-based Token Mixer for modeling nonlinear feature interactions within the MetaFormer framework. To utilize this nonlinear modeling capability while maintaining computational efficiency, we propose a dual-path Token Mixer architecture. One branch employs Separable Convolutions to efficiently capture local patterns, while the other uses the proposed Polynomial Kernel Mixer in shallow stages and attention mechanisms in deeper stages to model feature interactions and global context. This design enables complementary feature extraction with limited computational overhead.
Experiments on ImageNet-1K demonstrated that the proposed architecture achieves a competitive accuracy-efficiency trade-off across multiple model scales. The results suggest that combining nonlinear correlation modeling with efficient local feature extraction provides an effective and practical design strategy for MetaFormer-based image classification models.

\end{abstract}

\section{Introduction}
\label{sec:intro}

Existing image classification architectures have achieved strong performance by effectively capturing local details and global context. 
To improve image classification performance, various architectures have been proposed, most notably Vision Transformers (ViTs) \cite{ViT} and the MetaFormer framework \cite{poolformer,mlpmmixer}. MetaFormer-based models, such as PoolFormer \cite{poolformer}, ConvFormer, and CAFormer \cite{metaformerbaselien}, have demonstrated superior performance over conventional models like DeiT \cite{deit} and Swin Transformer \cite{swin} by optimizing the Token Mixer component. Although these recent architectures have achieved remarkable progress in capturing local and global representations, they often still insufficiently exploit correlation-aware information among feature responses. As a result, important relational cues may be overlooked, which can limit classification performance in challenging scenarios. Moreover, these models still suffer from an inherent trade-off between accuracy and computational efficiency.

Lightweight variants\cite{mobilenet,sepconv3,mobilenetv3} often sacrifice accuracy for speed, while high-performance models\cite{maxvit} incur substantial computational costs.
Moreover, reliance on a single Token Mixer\cite{rewrite,mambaout} tends to bias feature representations and can further increase computational complexity. Therefore, designing architectures that improve accuracy while maintaining computational efficiency remains an important challenge. 
Moreover, many recent Token Mixer\cite{inceptionnext,mambaout,MogaNet,OverLock} designs rely on linear transformations combined with activation functions to introduce nonlinearity. Although these architectures have demonstrated strong performance, higher-order feature interactions are generally represented only implicitly through layered linear operations and nonlinear activations. This suggests that explicitly modeling higher-order feature interactions within the Token Mixer itself remains an underexplored design space, with the potential to improve representation diversity and feature modeling capability.

\begin{figure}[t]
    \centering
    \includegraphics[width=0.7\linewidth]{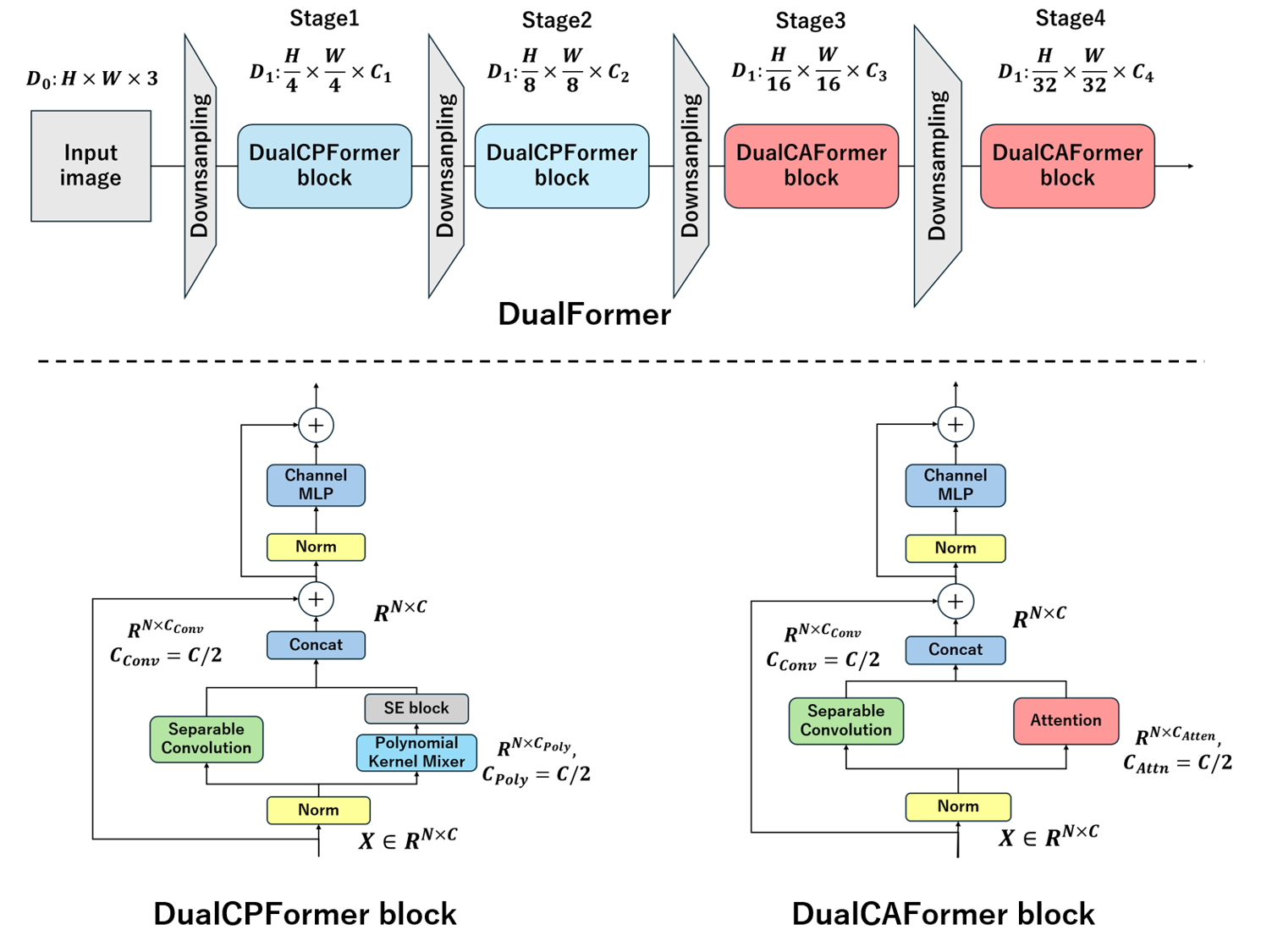}
    \caption{Overview of DualFormer. Bottom-left shows DualCPFormer employing the Dual Token Mixer, and the bottom-right shows DualCAFormer employing the Dual Token Mixer.}
    \label{dualformer_overview}
\end{figure}

In this paper, we focus on the representational limitations imposed by a single Token Mixer and propose a Dual Token Mixer that splits the input features into two branches, applies different Token Mixers to each branch, and then integrates them.
As shown in \cref{dualformer_overview}, 
one branch employs Separable Convolution \cite{sepconv1,sepconv2,sepconv3}, which effectively captures local features while preserving computational efficiency. The other branch introduces a Polynomial Kernel Mixer based on the Polynomial Kernel \cite{polynomialkernel} to explicitly model higher-order dependencies among feature responses, thereby enhancing correlation-aware representations.
The Polynomial Kernel Mixer is applied in the shallow stages where the input resolution is high, enabling the modeling of complex nonlinear feature interactions from rich spatial representations.

Since these early stages contain abundant fine-grained spatial information, 
explicitly modeling feature interactions at this stage can effectively enhance representational capacity. Therefore, we restrict the Polynomial Kernel Mixer to the shallow layers to fully exploit high-resolution features. Furthermore, as shown in \cref{dualformer_overview}, in the deeper stages where the spatial resolution becomes lower, Self-Attention can be used more efficiently, and we therefore employ standard Self-Attention to capture global contextual information.
By employing the Dual Token Mixer as its core component, the proposed DualFormer integrates local features through Separable Convolution, global contextual information through Self-Attention, and feature interactions through the Polynomial Kernel Mixer. This complementary design enables the acquisition of diverse feature representations while maintaining computational efficiency.

\begin{figure}[t]
    \centering

    \begin{minipage}[b]{0.45\linewidth}
        \centering
        \includegraphics[width=\linewidth]{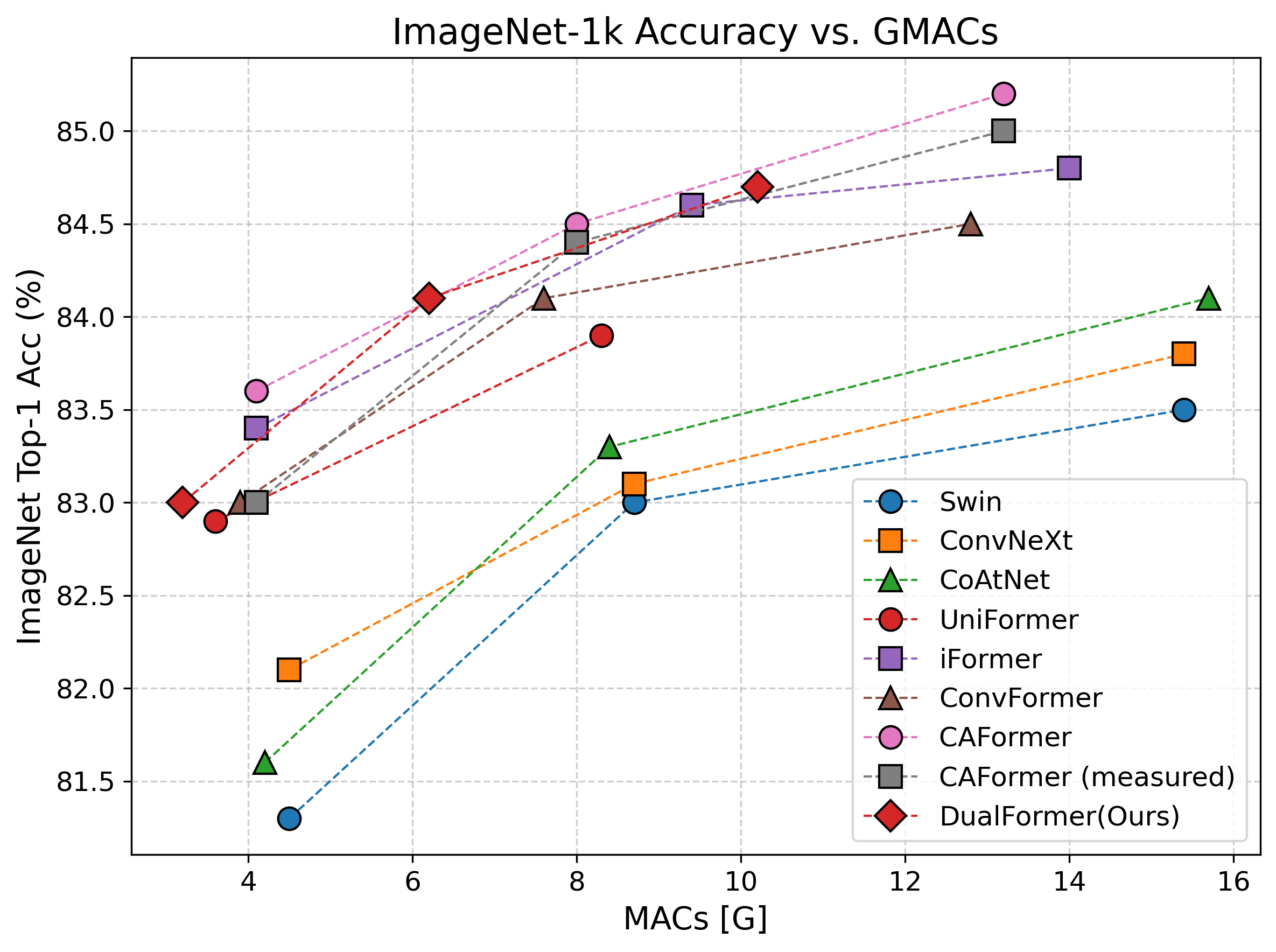}
    \end{minipage}
    \hfill
    \begin{minipage}[b]{0.45\linewidth}
        \centering
        \includegraphics[width=\linewidth]{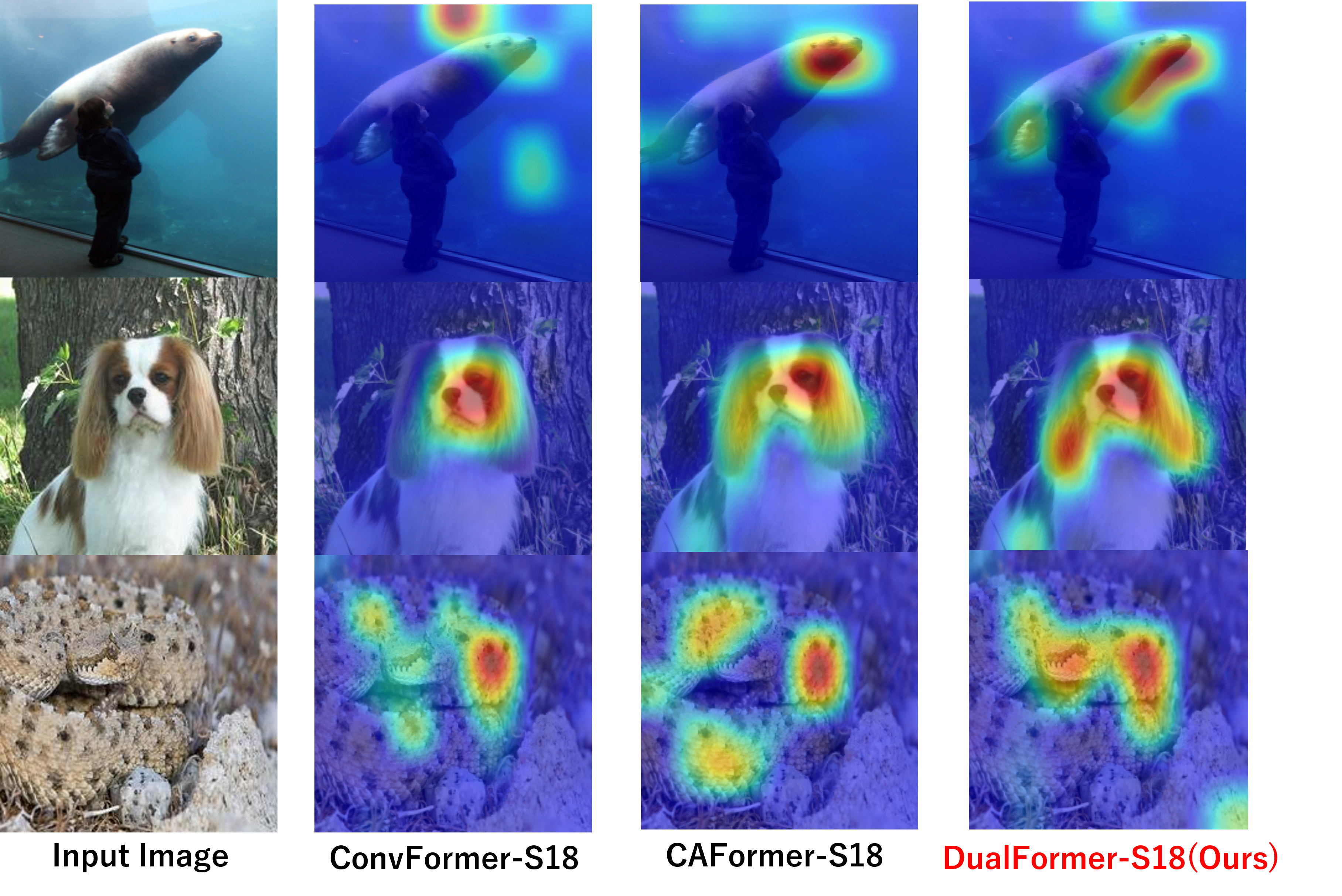}
    \end{minipage}

    \caption{
    Left: Accuracy-efficiency comparison on ImageNet-1K.
    Right: Grad-CAM visualizations of ConvFormer-S18, CAFormer-S18, and DualFormer-S18. DualFormer attends to broader object regions while preserving discriminative local details.
    }
    \label{fig:result}
\end{figure}

Evaluations on ImageNet-1K\cite{imagenet} demonstrate that DualFormer achieves a competitive accuracy-efficiency trade-off across multiple model scales, as shown in \cref{fig:result}. Compared with PoolFormerV2-S12, DualFormer-S18 improves Top-1 accuracy from 78.0\% to 83.0\%. DualFormer-S18 also achieves 1.7\% higher Top-1 accuracy than Swin-T while reducing MACs by approximately 29\%. Furthermore, DualFormer-S18 achieves the same Top-1 accuracy as ConvFormer-S18 while reducing MACs by approximately 18\%. Compared with CAFormer-S18, DualFormer-S18 matches its measured Top-1 accuracy while reducing MACs by approximately 22\%. These results highlight the effectiveness of combining explicit correlation modeling with efficient feature extraction.

To investigate the effectiveness of the proposed design, we visualize feature maps by Grad-CAM \cite{GradCAM} in \cref{fig:result}. The visualizations suggest that DualFormer learns representations that capture local patterns, global context, and feature interactions in a complementary manner.

\section{Previous Token Mixers}
\label{sec:realetd}

Conventional token mixers \cite{poolformer,metaformerbaselien,transnext} are primarily based on sequential designs, where a single module is applied to each stage of the network. In particular, hierarchical designs that switch between different token mixers according to network depth are widely adopted.

As a representative example, ConvFormer \cite{metaformerbaselien} employs Separable Convolution as its token mixer, enabling consistent extraction of local features across all stages and thereby contributing to performance improvement. In contrast, CAFormer \cite{metaformerbaselien} applies Separable Convolution to the early stages (Stages 1 and 2) and switches to an attention-based token mixer in the later stages, thereby achieving a balance between local and global feature representations. Furthermore, TransNeXt \cite{transnext} introduces different attention mechanisms depending on network depth, aiming to balance efficiency and global modeling capability.

However, these sequential designs have several limitations. Separable Convolution suffers from a computational bottleneck caused by pointwise convolution, and attention-based modules generally incur high computational cost. In addition, when attention is introduced only in later stages, the modeling of long-range dependencies in early stages becomes limited, which may lead to representations biased toward local inductive priors.

In contrast, alternative approaches have been proposed that do not rely on a single token mixer but instead combine multiple receptive fields or heterogeneous feature extraction mechanisms in parallel. For example, \cite{googlnet,resnextCVPR} achieve diverse receptive field modeling by constructing multiple convolutional paths in parallel. Similarly, \cite{inceptionnext} splits input features and applies convolutions with different kernel sizes in parallel before aggregating them, enabling efficient multi-scale representation learning. Furthermore, \cite{caretransformer} divides features and applies different token mixers (Separable Convolution and Linear Attention) to each branch, thereby integrating local and global feature representations.

On the other hand, these methods aim to improve representational capacity by combining multiple receptive fields or heterogeneous feature extraction pathways. However, the feature transformations within each pathway are often realized through combinations of convolutional operations, linear projections, or sequential nonlinear transformations. 
As a result, higher-order interactions among features may be captured only implicitly through successive transformations, rather than being explicitly modeled as a dedicated architectural component. This implicit formulation can limit the ability of the network to effectively emphasize complex nonlinear dependencies among features.

From this observation, it is considered that explicitly introducing higher-order feature interactions within the token mixer may improve representation efficiency while keeping additional computational overhead limited.
Based on this hypothesis, we follow a feature-splitting strategy similar to \cite{caretransformer}, while proposing DualFormer (\cref{dualformer_overview}) that introduces nonlinear feature interactions based on a Polynomial Kernel, with the aim of jointly capturing local features, global features, and higher-order feature interactions.

\section{Proposed Method}
\label{sec:method}
\label{dualca}

\subsection{MetaFormer}

MetaFormer is not a specific model but a general architecture designed to unify various vision models by not explicitly specifying the token mixer. Given an input image \(I\), it is first transformed into a sequence of tokens through an embedding layer as
\begin{equation}
\small
X = \mathrm{InputEmbedding}(I),
\end{equation}
where \(X \in \mathbb{R}^{N \times C}\) denotes a token sequence with length \(N\) and channel dimension \(C\).

A MetaFormer block is formulated as
\begin{align}
\small
X' &= X + \mathrm{TokenMixer}(\mathrm{Norm}_1(X)), \\
X'' &= X' + \sigma(\mathrm{Norm}_2(X')W_1)W_2,
\end{align}
where \(\mathrm{TokenMixer}(\cdot)\) is a module responsible for information propagation across tokens, \(\mathrm{Norm}_1(\cdot)\) and \(\mathrm{Norm}_2(\cdot)\) are normalization layers, \(\sigma(\cdot)\) is a nonlinear activation function, and \(W_1, W_2\) are learnable parameters of the channel MLP.

Within this formulation, specifying a concrete token mixer yields specific models such as ConvFormer and CAFormer. A representative method, CAFormer, adopts a hierarchical design in which different token mixers are sequentially switched according to network depth. Specifically, shallow stages extract local features using convolution-based token mixers, while deeper stages model global dependencies using Self-Attention.

\subsection{DualFormer}

In this work, we propose a Dual Token Mixer (see \cref{dualformer_overview}) that applies multiple token mixers in parallel within a single stage, without relying on a single token mixer. Following \cite{caretransformer}, the input feature \(X\) is first split along the channel dimension as
\begin{equation}
\small
X_1, X_2 = \mathrm{Split}(\mathrm{Norm}_1(X)).
\end{equation}
The proposed method applies different token mixers to the split channel groups.

Based on this design, we instantiate two variants, namely \textbf{DualCAFormer} and \textbf{DualCPFormer}.
In \textbf{DualCAFormer}, the two branches are designed to capture local and global contextual information, respectively.
In contrast, \textbf{DualCPFormer} is designed such that its two branches capture local information and explicit second-order nonlinear information, respectively.
To fully exploit the characteristics of these two variants, DualFormer adaptively assigns them to different stages.
Specifically, in the shallow stages, i.e., Stages 1 and 2, where high-resolution feature maps dominate the computational cost, we adopt \textbf{DualCPFormer} for its efficiency and strong nonlinear modeling capability.
In the deeper stages, i.e., Stages 3 and 4, we adopt \textbf{DualCAFormer} to jointly model local and global features via self-attention.
Using a single type of block for all stages results in an inherent trade-off between computational efficiency and representation capacity.
By contrast, our stage-wise design integrates the complementary strengths of both variants, thereby achieving a balanced architecture in terms of efficiency and representational power.

\noindent \textbf{DualCAFormer}

DualCAFormer performs parallel extraction of local and global features as
\begin{align}
\small
X' &= X + \mathrm{Concat}\big(
\mathrm{SelfAttention}(X_1),
\mathrm{SeparableConv}(X_2)
\big), \\
X'' &= X' + \sigma(\mathrm{Norm}_2(X')W_1)W_2,
\label{eqdualformer}
\end{align}
which enables simultaneous modeling of local inductive biases and global dependencies within a single stage.

Unlike \cite{caretransformer}, which employs Linear Attention~\cite{linearatteniton1,linearatteniton2} together with Separable Convolution, the proposed DualCAFormer adopts Self-Attention and Separable Convolution.
In general, the computational complexities of Self-Attention and Linear Attention are given by
\begin{equation}
\small
\mathcal{O}_{\mathrm{SA}} = \mathcal{O}(N^2C), \quad
\mathcal{O}_{\mathrm{LA}} = \mathcal{O}(NC^2),
\end{equation}
where $N$ denotes the number of tokens and $C$ denotes the feature dimension.

In Stage 3 and Stage 4, where DualCAFormer is applied, the spatial resolution is reduced, resulting in
$(N, C) = (196, 320)$ and $(49, 512)$, respectively.
Substituting these values into the complexity expressions yields
$\mathcal{O}_{\mathrm{SA}} = 1.23\times10^7,\ 1.23\times10^6, 
\mathcal{O}_{\mathrm{LA}} = 2.01\times10^7,\ 1.28\times10^7$
for Stage 3 and Stage 4 respectively, 
indicating that Self-Attention is computationally more efficient than Linear Attention in these stages.

Consequently, the computational overhead of Self-Attention remains limited in the later stages. 
Furthermore, Self-Attention directly models interactions among all tokens without approximation.
Based on this observation, Self-Attention is adopted in DualCAFormer to capture global feature dependencies while maintaining computational efficiency.

\noindent \textbf{DualCPFormer}

In DualCPFormer of \cref{dualformer_overview}, Self-Attention is replaced with a Polynomial Kernel Mixer, 
which is a polynomial-kernel-based token-mixing module designed to explicitly model nonlinear and higher-order feature interactions.
Although its detailed formulation is provided in the following subsection, 
we first define the overall computation of the DualCPFormer block as
\begin{align}
\small
Y_1 &= \mathrm{SeparableConv}(X_1), \quad
\tilde{Y}_2 = \mathrm{PolynomialKernelMixer}(X_2), \\
Y_2 &= \mathrm{SE}(\tilde{Y}_2), \\
X' &= X + \mathrm{Concat}(Y_1, Y_2), \\
X'' &= X' + \sigma(\mathrm{Norm}_2(X')W_1)W_2.
\label{dualCP}
\end{align}
where $\mathrm{SeparableConv}(\cdot)$ denotes a depthwise separable convolution for capturing local information, 
whereas $\mathrm{PolynomialKernelMixer}(\cdot)$ denotes the proposed polynomial-kernel-based token mixer described later.
$\mathrm{SE}(\cdot)$ represents the squeeze-and-excitation module for adaptive channel-wise reweighting, 
and $\mathrm{Concat}(\cdot)$ denotes channel-wise concatenation.
Furthermore, $W_1$ and $W_2$ are learnable linear projection matrices, 
and $\sigma(\cdot)$ denotes a nonlinear activation function.

\noindent \textbf{A Novel Token Mixer Using a Polynomial Kernel}

\label{dualcp}
\begin{figure*}[t]
    \centering
    \includegraphics[width=\textwidth]{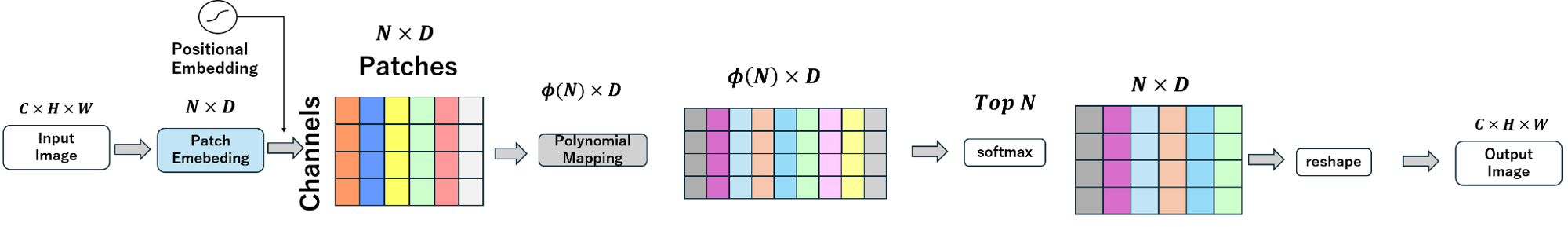}
    \caption{Overview of the Polynomial Kernel Mixer}
    \label{polynomialkernelmixer}
\end{figure*}

In shallow stages, the input resolution is high, and directly applying Self-Attention becomes computationally dominant, leading to a significant efficiency constraint. On the other hand, in visual recognition, it is important not only to model local information and global dependencies, but also to handle interactions between them within a unified representation space. 

However, existing token mixers often rely on convolution operations or linear weighted aggregation in attention mechanisms, where nonlinear and higher-order interactions between local and global features are expected to emerge only implicitly. As a result, mechanisms for explicitly modeling and selectively enhancing such interactions are not sufficiently incorporated into the architecture.

In this work, we introduce a new token mixer, the Polynomial Kernel Mixer shown in \cref{polynomialkernelmixer}, which aims to satisfy computational efficiency constraints while explicitly modeling input-dependent feature interactions.
In the Polynomial Kernel Mixer, the input feature $X \in \mathbb{R}^{H \times W \times C}$ is first divided into patches of size $P \times P$ and transformed into token embeddings as
\begin{equation}
\small
X_p = \mathrm{PatchEmbed}(X), \quad X_p \in \mathbb{R}^{N \times D}, \quad N = \frac{HW}{P^2}.
\end{equation}

Following ViT, a learnable positional embedding is added as
\begin{equation}
\small
X_0 = X_p + E_{\mathrm{pos}}, \quad E_{\mathrm{pos}} \in \mathbb{R}^{N \times D}.
\end{equation}

Then, a mapping based on a Polynomial Kernel is applied along the token dimension as
\begin{equation}
\small
\tilde{X} = \Phi(X_0).
\end{equation}
where $\Phi(\cdot)$ denotes a polynomial expansion, and $r$ represents the polynomial order.
In this study, we set $r=2$, corresponding to a second-order polynomial expansion.
For example, for a two-dimensional input $\mathbf{x} = [x_1, x_2]$, it is defined in \cref{phi}. In general, when the input dimension is $N$, the dimensionality after expansion is given in \cref{phiryou}. However, since both computational cost and memory usage increase rapidly for $r>2$, we restrict the setting to $r=2$.

\begin{equation}
\label{phi}
\small
\Phi(\tilde{\mathbf{X}}) = \left( x_1^2, \sqrt{2}x_1x_2, x_2^2 \right),
\end{equation}
\begin{equation}
\label{phiryou}
\small
\dim(\Phi(\mathbf{x})) = N_{\text{ST}} + \frac{N(N-1)}{2}_{\text{CT}}.
\end{equation}
where \(N_{ST}\) and \(N_{CT}\) denote the squared and cross terms.
This mapping produces feature representations that include both squared terms and cross terms, thereby explicitly introducing combinatorial interactions between input tokens.

Furthermore, softmax normalization is applied to the expanded representation along the token dimension to compute the relative importance of each component as
$\mathbf{A} = \mathrm{Softmax}_{N}(\tilde{X})$.
Then, the indices of the top \(N\) responses are selected based on the normalized scores as
$\mathcal{I} = \mathrm{TopN}(\mathbf{A})$.
By using the indices, the corresponding components are gathered from the expanded representation and reshaped back to original spatial feature shape as
\begin{eqnarray}
\label{select_phi}
\small
\mathbf{Z} &=& \mathrm{Reshape}\left(\mathrm{Gather}(\tilde{X}, \mathcal{I})\right),
\end{eqnarray}
where \(\mathbf{Z} \in \mathbb{R}^{H \times W \times C}\). 
The number of selected components is set to the number of patches \(N\), enabling the selected representation to be reshaped back to the original spatial resolution. This operation enables nonlinear feature transformation along the token dimension without increasing computational cost.

An analysis on ImageNet-1k using the S18 model variant shows that cross and squared terms are selected at rates of 73.18\% and 26.82\%, respectively. Although CT terms are selected more frequently, ST terms still constitute a non-negligible portion of the selected components. This suggests that the proposed mapping primarily captures cross-token interactions while retaining self-related nonlinear components, rather than collapsing into a single type of term.

In this work, the Polynomial Kernel Mixer is incorporated into the DualCPFormer block. The input features are split into two branches, where Separable Convolution is applied to one branch and the Polynomial Kernel Mixer is applied to the other, forming a structure that simultaneously handles local information and nonlinear feature transformations. As shown in \cref{dualformer_overview}, the output of the Polynomial Kernel Mixer is further refined by an SE block \cite{SEblock}, which performs channel-wise reweighting to enhance representation quality.

\section{Experimental Evaluation}
\label{sec:jikken}

\subsection{Experiment Setup}
In this study, we evaluate the performance of the proposed models on the image classification task using ImageNet-1K~\cite{imagenet}, a large-scale benchmark dataset. ImageNet-1K is one of the most widely used datasets in the field of computer vision, consisting of approximately 1.3 million training images from 1,000 classes and 50,000 validation images.The experimental settings follow those of ConvFormer in MetaFormer Baselines for Vision~\cite{metaformerbaselien}.

Although~\cite{metaformerbaselien} conducts training with a batch size of 4096, due to GPU memory constraints, we set the batch size to 1536 and the learning rate to $1.5 \times 10^{-3}$. The implementation is based on the PyTorch library~\cite{pytorch} and the Timm codebase~\cite{timm}. All experiments are conducted using three NVIDIA RTX A6000 GPUs.

To comprehensively evaluate the robustness of the models, we use several additional test sets in addition to ImageNet-1K (IN-1K). These include ImageNet-A (IN-A)~\cite{IMA}, which consists of naturally occurring adversarial examples; ImageNet-R (IN-R)~\cite{IMr}, an extended test set containing examples that are difficult for ResNet-50~\cite{RN50} to classify correctly; ImageNet-Sketch (IN-S)~\cite{IMsketch}, which consists of hand-drawn sketch images; and ImageNet-V2 (IN-V2)~\cite{IMv2}, an extended test set constructed using the same sampling strategy as ImageNet-1K.

As evaluation metrics, we use Top-1 accuracy (\%) to measure classification performance and multiply--accumulate operations (MACs, in giga scale) to measure computational cost. MACs reflect the total number of multiplication and addition operations required by a model.

Furthermore, to demonstrate that the proposed method is not limited to a single width--depth configuration, we implement DualFormer with multiple model sizes. The detailed configurations are shown in \cref{tab:dualformer_config}.

\begin{table}[h]
\centering
\caption{Architectural configurations of DualFormer.}
\label{tab:dualformer_config}
\scriptsize
\resizebox{0.5\linewidth}{!}{
\begin{tabular}{l|c|l}
\toprule
\multicolumn{2}{l|}{Model} & DualFormer (Ours) \\
\midrule
     & S18 & $C = (64, 128, 320, 512), \quad L = (3, 3, 9, 3)$ \\
     & S36 & $C = (64, 128, 320, 512), \quad L = (3, 12, 18, 3)$ \\
Size & M36 & $C = (96, 192, 384, 576), \quad L = (3, 12, 18, 3)$ \\
\midrule
\multicolumn{2}{l|}{Classifier Head} & Global average pooling, Norm, MLP \\
\bottomrule
\end{tabular}
}
\end{table}

\subsection{Experiment Results}

\begin{table}[t]
\centering
\caption{
The horizontal lines in the table indicate groups of models with similar GMACs ranges. Each row corresponds to a model, and each column reports the accuracy on a specific dataset: IN-1K, IN-A, IN-R, IN-S, and IN-V2. The IN-1K results are either taken from the corresponding papers or measured by ourselves; for CAFormer, both the reported results and our reproduced results are presented. The results on IN-A, IN-R, IN-S, and IN-V2 are evaluated by ourselves using only models with IN-1K pretrained weights available in the Timm library or publicly released implementations such as official GitHub repositories.}
\scriptsize
\resizebox{\linewidth}{!}{
\begin{tabular}{lcccccccc}
\toprule
Model & Type & Params (M) & MACs (G) & IN-1K,Top-1 (\%) & IN-A,Top-1 (\%) & IN-R,Top-1 (\%) & IN-S,Top-1 (\%) & IN-V2,Top-1 (\%) \\
\midrule
RSB-ResNet-18\cite{resbresnet1},\cite{resbresnet2}& CNN & 11.7 & 1.8 & 70.6 & 3.7 & 34.2 & 23.0 &61.8\\
IdentityFormer-S12\cite{metaformerbaselien}& MetaFormer & 11.9 & 1.8 & 74.6 & 5.1 & 32.8 & 21.0 & 62.0\\
RandFormer-S12\cite{metaformerbaselien}& MetaFormer & 11.9 + 0.2 & 1.9 & 76.6 & 9.2 & 35.2 & 23.7 & 64.5\\
PoolFormerV2-S12\cite{metaformerbaselien}& MetaFormer & 11.9 & 1.8 & 78.0 &8.9 & 39.2 & 27.6 & 66.2\\
BiFormer-T\cite{biformer}& MetaFormer-Hybrid & 13.1 & 2.2 & 81.4 & 25.6 & 45.3 & 31.5 & 70.6\\
CARE-S2\cite{caretransformer}& Hybrid & 19.5 & 1.9 & 82.1 & 23.0 & 47.8 & 36.2 & 71.7\\
iFormer-S\cite{iformer}& MetaFormer & 20 & 4.8 & 83.4 & 37.4 & 46.2 & 34.8 & 73.0\\
IdentityFormer-S24\cite{metaformerbaselien}& MetaFormer & 21.3 & 3.4 & 78.2 & 10.0 & 36.7 & 24.3 & 66.2\\
PoolFormerV2-S24\cite{metaformerbaselien}& MetaFormer & 21.3 & 3.4 & 80.7 & 17.1 & 42.7 & 31.4 & 69.3\\
RSB-ResNet-34\cite{resbresnet1},\cite{resbresnet2}& CNN & 21.8 & 3.7 & 75.5 & 10.5 & 40.3 & 29.0 & 67.2\\
RandFormer-S24\cite{metaformerbaselien}& MetaFormer & 21.3 + 0.5 & 3.5 & 78.8 & 25.9 & 37.5 & 25.9 & 67.1\\
DeiT-S\cite{deit}& MetaFormer & 22 & 4.6 & 79.8 & 18.7 & 42.2 & 29.4 & 68.5\\
UniFormer-S\cite{uniformer}& MetaFormer-Hybrid & 22 & 3.6 & 82.9 & 33.2 & 47.2 & 35.1 & 72.4\\
CoAtNet-0\cite{cotanet}& MetaFormer & 25 & 4.2 & 81.6 & 26.9 & 42.8 & 31.3 & 71.2\\
BiFormer-S\cite{biformer}& MetaFormer-Hybrid & 25.5 & 4.5 & 83.8 & 39.3 & 49.6 & 36.4 & 73.7 \\
RSB-ResNet-50\cite{resbresnet1},\cite{resbresnet2}& CNN & 26 & 4.1 & 79.8 & 17.6 & 41.8 & 30.4 & 70.5 \\
InceptionNeXt-T\cite{inceptionnext}&MetaFormer-Hybrid& 28.0 & 4.2 & 82.3 & 25.7 & 47.0 & 33.6 & 71.9\\
ConvNeXt-T\cite{Convnext}& MetaFormer-Hybrid & 29 & 4.5 & 82.1 & 24.0 & 47.1 & 33.8 & 71.0\\
Swin-T\cite{swin}& MetaFormer & 29 & 4.5 & 81.3 & 21.0 & 41.4 & 29.3 & 69.6\\
ConvFormer-S18\cite{metaformerbaselien}& MetaFormer & 27 & 3.9 & 83.0 & 28.4 & 48.0 & 34.2 & 72.8\\
CAFormer-S18\cite{metaformerbaselien}& MetaFormer & 26 & 4.1 & 83.6,83.0(measured) & 36.3 & 48.0 & 35.5 & 73.2\\
DualFormer-S18 (Ours)& MetaFormer & 23 & 3.2 & 83.0 & 31.5 & 47.2 & 34.5 & 72.5\\
\midrule
IdentityFormer-S36\cite{metaformerbaselien}& MetaFormer & 30.8 & 5.0 & 79.3 & 12.2 & 36.7 & 24.3 & 66.2\\
RSB-ResNet-101\cite{resbresnet1},\cite{resbresnet2}& CNN & 45 & 7.9 & 81.3 & 28.6 & 46.9 & 34.8 & 72.7\\
RandFormer-S36\cite{metaformerbaselien}& MetaFormer & 30.8 + 0.7 & 5.2 & 79.5 & 14.1 & 39.1 & 27.3 & 67.1\\
PoolFormerV2-S36\cite{metaformerbaselien}& MetaFormer & 30.8 & 5.0 & 81.6 & 19.3 & 45.2 & 32.8 & 70.6\\
ConvNeXt-S\cite{Convnext}& MetaFormer-Hybrid & 50 & 8.7 & 83.1 & 31.2 & 49.5 & 37.1 & 72.4\\
Swin-S\cite{swin}& MetaFormer & 50 & 8.7 & 83.0 & 32.3 & 45.1 & 32.4 & 72.1\\
CoAtNet-1\cite{cotanet}& MetaFormer & 42 & 8.4 & 83.3 & 35.5 & 46.4 & 34.9 & 72.8\\
UniFormer-B\cite{uniformer}& MetaFormer-Hybrid & 50 & 8.3 & 83.9 & - & - & - & -\\
InceptionNeXt-S\cite{inceptionnext}& MetaFormer-Hybrid & 49 & 8.4 & 83.5 & 33.3 & 50.3 & 36.9 & 73.3\\
MogaNet-S\cite{MogaNet}& Hybrid & 25.3 & 5.0 & 83.4 & 13.9 & 40.4 & 28.7 & 68.0\\
TransNeXt-Tiny\cite{transnext}& MetaFormer-Hybrid & 28.2 & 5.7 & 84.0 & 39.7 & 49.6 & 37.6 & 73.8\\
ConvFormer-S36\cite{metaformerbaselien}& MetaFormer & 40 & 7.6 & 84.1 & 36.5 & 50.0 & 37.5 & 73.9\\
CAFormer-S36\cite{metaformerbaselien}& MetaFormer & 39 & 8.0 & 84.5,84.4(measured) & 44.0 & 50.5 & 38.6 & 74.8\\
DualFormer-S36 (Ours)& MetaFormer & 34 & 6.2 & 84.1 & 38.2 & 50.1 & 39.3 & 73.6\\
OverLoCK-T\cite{OverLock}& Hybrid & 33 & 5.5 & 84.2 & 40.6 & 53.8 & 41.0 & 74.1\\
\midrule
RSB-ResNet-152\cite{resbresnet1},\cite{resbresnet2}& CNN & 60 & 11.6 & 81.8 & 33.7 & 49.3 & 37.9 & 73.4\\
BiFormer-B\cite{biformer}& MetaFormer-Hybrid & 56.8 & 9.8 & 84.3 & 44.0 & 49.6 & 35.3 & 74.0\\
ConvNeXt-B\cite{Convnext}& MetaFormer-Hybrid & 89 & 15.4 & 83.8 & 36.5 & 51.3 & 38.2 & 73.7\\
DeiT-B\cite{deit}& MetaFormer & 86 & 17.5 & 81.8 & 27.1 & 44.9 & 32.4 & 71.2\\
Swin-B\cite{swin}& MetaFormer & 88 & 15.4 & 83.5 & 35.6 & 46.5 & 32.6 & 72.4\\
InceptionNeXt-B\cite{inceptionnext}& MetaFormer-Hybrid& 87 & 14.9 & 84.0 & 36.4 & 51.6 & 38.5 & 74.4\\
CoAtNet-2\cite{cotanet}& MetaFormer & 75 & 15.7 & 84.1 & - & - & - & -\\
iFormer-B\cite{iformer}& MetaFormer & 48 & 9.4 & 84.6 & 45.4 & 51.0 & 37.8 & 74.5\\
ConvFormer-M36\cite{metaformerbaselien}& MetaFormer & 57 & 12.8 & 84.5 & 41.1 & 50.1 & 37.9 & 74.4\\
iFormer-L\cite{iformer}& MetaFormer & 87 & 14.0 & 84.8 & 45.8 & 49.9 & 37.8 & 75.1\\
TransNeXt-small\cite{transnext}& MetaFormer-Hybrid & 49.7 & 10.3 & 84.7 & 46.5 & 52.5 & 39.7 & 74.7\\
MogaNet-B\cite{MogaNet}& Hybrid & 43.9 & 9.9 & 84.3 & 40.6 & 50.0 & 38.1 & 74.0\\
TransNeXt-base\cite{transnext}& MetaFormer-Hybrid& 89.7 & 18.4 & 84.8 & 50.0 & 53.9 & 41.4 & 75.1\\
CAFormer-M36\cite{metaformerbaselien}& MetaFormer & 56 & 13.2 & 85.2,85.0(measured) & 49.1 & 50.7 & 38.5 & 75.6\\
DualFormer-M36 (Ours)& MetaFormer & 49 & 10.2 & 84.7 & 43.6 & 49.9 & 37.6 & 74.3\\
OverLoCK-S\cite{OverLock}& Hybrid & 56 & 9.7 & 84.8 & 45.8 & 57.1 & 45.7 & 74.8\\
\midrule
\bottomrule
\end{tabular}
}
\label{jikkennkekka}
\end{table}

The experimental results are presented in \cref{jikkennkekka}. 
The ``Type'' column categorizes models using CNN blocks as ``CNN,'' pure MetaFormer-based models composed of a Token Mixer and an FFN as ``MetaFormer,'' models with structural modifications based on MetaFormer as ``MetaFormer-Hybrid,'' and models incorporating additional modules such as memory or gating mechanisms as ``Hybrid.''

As shown in \cref{jikkennkekka}, DualFormer achieves competitive performance with fewer parameters and lower computational cost than existing MetaFormer-based models across different model sizes. 
In particular, DualFormer-S18 achieves 83.0\% Top-1 accuracy on IN-1K with 23M parameters and 3.2 GMACs. 
This is comparable to ConvFormer-S18 while reducing the number of parameters from 27M to 23M and the MACs from 3.9G to 3.2G. 
Furthermore, DualFormer-S18 achieves the same IN-1K accuracy as our reproduced CAFormer-S18 with lower computational cost.

These results indicate that DualFormer effectively integrates two feature extraction paths, DualCPFormer and DualCAFormer, without relying on a single Token Mixer. 
By modeling local and global features in parallel, DualFormer achieves high accuracy with reduced computational cost. 
In contrast, conventional sequential Token Mixers require more computation to obtain comparable accuracy, highlighting the favorable balance between accuracy and efficiency achieved by DualFormer.

On the other hand, DualFormer-S18 underperforms CAFormer-S18 on IN-A, IN-R, and IN-V2, while maintaining relatively competitive performance on IN-S. 
In particular, DualFormer-S18 achieves 31.5\% on IN-A, whereas CAFormer-S18 achieves 36.3\%. 
This suggests that, although DualFormer-S18 is computationally efficient, it may be less robust to challenging out-of-distribution or adversarial samples such as IN-A. 
Interestingly, the performance gap on IN-S is relatively small, possibly because the local feature extraction and explicit nonlinear feature interactions in DualCPFormer are effective for capturing contour- and edge-based representations.

A similar tendency is observed for medium-scale models. 
DualFormer-S36 achieves 84.1\% Top-1 accuracy on IN-1K with 34M parameters and 6.2 GMACs, matching ConvFormer-S36 while reducing both parameters and MACs. 
It also achieves 50.1\% and 39.3\% on IN-R and IN-S, respectively, slightly outperforming ConvFormer-S36. 
In particular, DualFormer-S36 surpasses CAFormer-S36 on IN-S (38.6\%), suggesting its effectiveness for inputs with large variations in shape and texture. 
However, compared with CAFormer-S36, it still shows performance gaps on IN-1K, IN-A, IN-R, and IN-V2, especially on IN-A, where DualFormer-S36 achieves 38.2\% while CAFormer-S36 achieves 44.0\%. 
Furthermore, the Hybrid model OverLoCK-T outperforms DualFormer-S36 across all robustness benchmarks despite having a similar parameter scale, suggesting that gating-based Hybrid models may be advantageous for handling out-of-distribution data.

For the large-scale model, DualFormer-M36 achieves 84.7\% Top-1 accuracy on IN-1K with 49M parameters and 10.2 GMACs, outperforming ConvFormer-M36 while reducing both parameters and computational cost. 
It also outperforms MogaNet-B on IN-1K, IN-A, and IN-V2, demonstrating strong efficiency while preserving the core MetaFormer structure.

However, DualFormer-M36 still underperforms CAFormer-M36 on IN-1K, IN-A, and IN-V2. 
In addition, MetaFormer-Hybrid and Hybrid models such as TransNeXt and OverLoCK achieve higher robustness performance. 
In particular, OverLoCK-S significantly outperforms DualFormer-M36 on IN-R and IN-S. 
Therefore, although DualFormer-M36 provides a favorable balance between efficiency and accuracy, it does not achieve the best performance on out-of-distribution benchmarks compared with more complex Hybrid models.

Overall, DualFormer achieves high classification performance and computational efficiency across a wide range of model scales while maintaining a pure MetaFormer-based architecture without additional external modules, as shown in \cref{fig:result}. 
For small- and medium-scale models, it achieves comparable or better performance than existing MetaFormer-based models with fewer parameters and lower MACs. 
For larger models, the performance gap with CAFormer becomes smaller, while Hybrid models tend to outperform DualFormer on some robustness benchmarks. 
Therefore, the main advantage of DualFormer lies not in achieving the highest accuracy, but in realizing a favorable trade-off among accuracy, robustness, model size, and computational cost while preserving a pure MetaFormer-based architecture.

\subsection{Ablation Study}

\begin{figure}[t]
    \centering
    \includegraphics[width=0.8\linewidth]{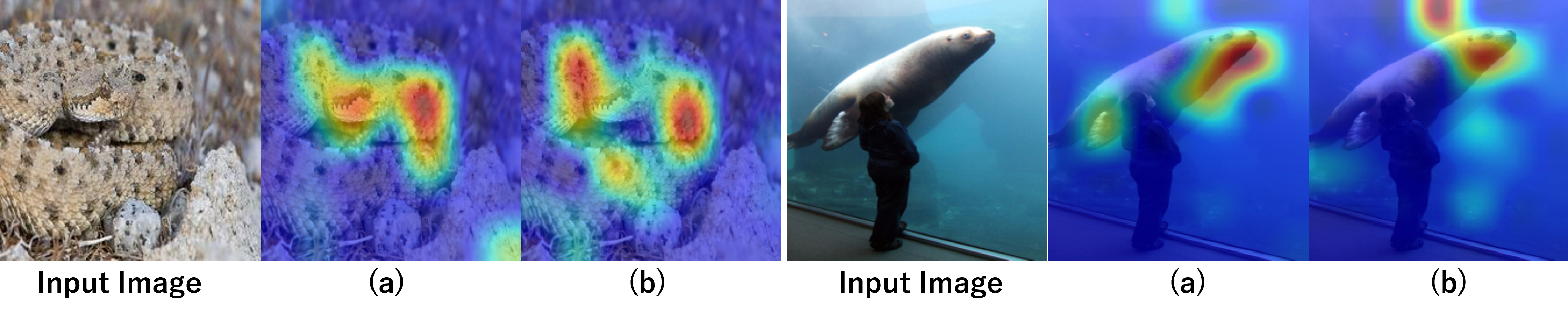}
    \vspace{-2mm}
    \caption{Grad-CAM at the final stage in DualFormer-S18. Left (a): proposed method. Right (b): Polynomial Kernel Mixer and SE block replaced with identity mappings.}
    \label{gradcamablation}
    \vspace{-2mm}
\end{figure}

\begin{table}[t]
\centering
\vspace{-2mm}
\caption{Ablation study of the Polynomial Kernel Mixer in DualCPFormer.}
\label{ablation}
\resizebox{\linewidth}{!}{
\begin{tabular}{lcccccccc}
\toprule
Model & Polynomial Kernel Mixer \& SE Block & Params (M) & MACs (G) & IN-1K Top-1 (\%) & IN-A Top-1 (\%) & IN-R Top-1 (\%) & IN-S Top-1 (\%) & IN-V2 Top-1 (\%)\\
\midrule
DualFormer-S18 (Ours) & \checkmark & 23 & 3.2 & \textbf{83.0} & \textbf{31.5} & \textbf{47.2} & \textbf{34.5} & \textbf{72.5} \\
DualFormer-S18 &  & 23 & 3.2 & 82.8 & 29.9 & 46.3 & 33.3 & 72.2\\
\midrule
DualFormer-S36 (Ours) & \checkmark & 34 & 6.2 & \textbf{84.1} & 38.2 & \textbf{50.1} & \textbf{39.3} & 73.6 \\
DualFormer-S36 &  & 34 & 6.2 & 83.9 & \textbf{39.3} & 49.8 & 37.8 & \textbf{74.2}\\
\midrule
DualFormer-M36 (Ours) & \checkmark & 49 & 10.2 & \textbf{84.7} & 43.6 & 49.9 & 37.6 & \textbf{74.3} \\
DualFormer-M36 &  & 49 & 10.2 & 84.6 & \textbf{44.3} & \textbf{50.4} & \textbf{38.1} & 74.2\\
\bottomrule
\end{tabular}
}
\vspace{-2mm}
\end{table}

In this section, we conduct an ablation study to evaluate the effectiveness of the proposed Polynomial Kernel Mixer and the SE Block in DualCPFormer by replacing both components with identity mappings. The results are shown in \cref{ablation}.
Incorporating the Polynomial Kernel Mixer and the SE Block improves the Top-1 accuracy on IN-1K across all model sizes, indicating that these components contribute to the standard classification performance of DualCPFormer. More importantly, the proposed components also improve robustness on several OOD benchmarks. 
For DualFormer-S18, consistent performance gains are observed on all evaluation sets, including IN-A, IN-R, IN-S, and IN-V2. For DualFormer-S36, improvements are also obtained on IN-R and IN-S, suggesting that the proposed components are effective in enhancing robustness to certain distribution shifts, such as rendition and sketch-style variations.
These results indicate that DualCPFormer benefits from the proposed nonlinear token mixing and channel recalibration mechanisms, particularly in small and medium model variants. On the other hand, as the model size increases, the improvements on OOD benchmarks become less consistent. For DualFormer-M36, the gains are mainly limited to IN-1K and IN-V2, while performance degradation is observed on IN-A, IN-R, and IN-S.

Furthermore, the Grad-CAM visualizations shown in \cref{gradcamablation} indicate that, when the Polynomial Kernel Mixer and the SE Block are removed, the model tends to focus mainly on contours and edge regions. In contrast, the proposed method attends to semantically meaningful regions, such as faces.

\section{Discussion}
\label{sec:dis}

\subsection{Design Analysis}
In this section, we discuss the design principles of DualCPFormer and DualCAFormer, as illustrated in \cref{dualformer_overview,polynomialkernelmixer}.

DualCAFormer is applied only to the later stages, namely Stages 3 and 4, where the spatial resolution is relatively small. 
The main reason for this design is that the computational cost of Self-Attention increases quadratically with respect to the input resolution. 
If Self-Attention is applied to earlier stages with larger spatial resolutions, the computational cost increases significantly, while the expected performance gain may be limited. 
From the perspective of feature representation, early stages are dominated by low-level spatial features such as edges and textures, whereas later stages contain more abstract semantic features. 
Since Self-Attention is effective at capturing long-range relationships among features, applying it to later stages, where semantic features are richer, is expected to provide benefits that justify its computational cost. 
Similar observations have also been reported in \cite{tinynext}.

For DualCPFormer, the patch size of the Polynomial Kernel Mixer is set to $14 \times 14$. 
This design choice is based on the trade-off between the computational cost of feature interactions induced by the Polynomial Kernel and the resulting representational capacity. 
In this study, we set the polynomial order to $r=2$. 
Increasing the polynomial order rapidly increases the number of generated terms. 
For example, when $r=3$, the number of terms becomes
\begin{equation}
\small
N + N(N-1) + \frac{N(N-1)(N-2)}{6},
\end{equation}
which may lead to increased memory consumption and a stronger selection bias in the Softmax operation. 
Therefore, we adopt $r=2$ to balance computational efficiency and representational capacity.

The choice of patch size is also based on the trade-off between the feature region to be captured and computational complexity. 
The Separable Convolution branch, which runs in parallel with the Polynomial Kernel Mixer, extracts local features with a receptive field of $7 \times 7$, as in ConvFormer and CAFormer. 
Therefore, if an equal or smaller patch size is used, the features extracted by the Polynomial Kernel are likely to be limited to local nonlinear correlations. 
To capture feature relationships over a wider region than Separable Convolution, we adopt a patch size of $14 \times 14$.

Furthermore, for an input size of $H=W=56$, the adopted patch size of $14 \times 14$ produces $N=16$ tokens, resulting in 136 generated polynomial terms. 
In contrast, when a smaller patch size of $7 \times 7$ is used, the number of tokens increases to $N=64$, and the number of generated terms significantly increases to 2,080. 
Therefore, the $14 \times 14$ patch size can be regarded as a practical design choice that suppresses computational and memory costs while preserving the ability to capture relatively wide-range feature interactions.

\subsection{Effect of Explicit Second-Order Interactions}

The results in \cref{ablation} suggest that the Polynomial Kernel Mixer may compensate for the limited representational capacity of small-scale models by explicitly introducing 
second-order feature interactions. It has also been reported in~\cite{polynomialnetworks,bilinearnet} that explicitly modeling 
of second-order feature interactions enables the efficient representation of complex feature relationships. Although deep neural networks can, in principle, implicitly learn relationships among higher-order features, sufficient model capacity is required to acquire such representations. In contrast, our method explicitly introduces higher-order interactions into the representation space, which allows the model to efficiently exploit complex feature relationships even under limited model capacity. In fact, the largest performance improvement is observed in the smallest model, DualFormer-S18, which is consistent with this interpretation.

On the other hand, as model capacity increases, the backbone itself becomes more capable of implicitly learning interactions among higher-order features. Therefore, the additional representational benefit provided by the Polynomial Kernel Mixer is expected to become relatively smaller. Moreover, while explicit second-order feature interactions can improve representational capacity, when the model capacity is sufficiently large, they may overlap with already learned feature representations. Such redundancy may introduce less useful features and does not necessarily lead to improved OOD performance. The limited improvement observed in DualFormer-M36 is consistent with this interpretation.

In addition, the consistent performance improvement on IN-S suggests that second-order feature interactions may contribute to represent structural characteristics of objects. Since ImageNet-Sketch is a benchmark in which texture information is largely removed and classification relies more heavily on object contours and structural cues, it is possible that second-order feature interactions enhance structural feature representations. However, since this study does not directly evaluate shape bias or the properties of learned feature representations, further investigation is required to clarify this relationship.

Furthermore, the Grad-CAM results confirm that our modules encourage the model to attend not only to simple contour information but also to semantically meaningful regions that are effective for discrimination. This suggests that correlations among features are effectively exploited through 
second-order feature interactions, resulting in the formation of more discriminative feature representations.
Overall, these results suggest that the Polynomial Kernel Mixer is particularly effective for small-scale models and plays a supportive role in learning higher-order feature representations. However, its effectiveness depends on model capacity, and its impact may become limited in large-scale models.

\section{Conclusion}
\label{sec:coclue}

In this study, we proposed DualFormer, a network architecture with dual feature extraction paths. Experimental results demonstrate that DualFormer achieves comparable or superior classification performance compared to ConvFormer and CAFormer, while significantly reducing both the number of parameters and computational cost. In particular, we confirm that the combination of DualCPFormer and DualCAFormer blocks enables efficient integration of local features, global features, and feature interactions.

Furthermore, DualFormer achieves comparable Top-1 accuracy per GMACs to existing Token Mixers, demonstrating its high computational efficiency. In addition, through ablation studies, we show that explicit feature interactions introduced by the Polynomial Kernel Mixer can compensate for the limited capacity of small-scale models, contributing to improvements in both classification performance and robustness. 
Overall, these results indicate that DualFormer not only achieves a favorable balance between accuracy and computational efficiency but also demonstrates the effectiveness of efficient feature representation learning through explicit higher-order feature interactions.

\bibliographystyle{splncs04}
\bibliography{main}

\clearpage
\appendix
\section{Ablation of the Components in the Polynomial Kernel Mixer}
\cref{tab:ablation_detail} presents the ablation results for the Polynomial Kernel Mixer and the SE Block.
The configuration combining both achieves the best or on-par-with-best results on nearly all metrics, including IN-1K, and outperforms the other configurations on IN-A and IN-R in particular.
We therefore adopt both Polynomial Kernel Mixer and the SE Block in our method.

\begin{table}[h]
\centering
\vspace{-2mm}
\caption{Ablation study of the Polynomial Kernel Mixer  and SE Block in DualFormer-S18.}
\label{tab:ablation_detail}
\resizebox{\linewidth}{!}{
\begin{tabular}{lccccccccc}
\toprule
Model & Polynomial Kernel Mixer & SE Block & Params (M) & MACs (G) & IN-1K Top-1 (\%) & IN-A Top-1 (\%) & IN-R Top-1 (\%) & IN-S Top-1 (\%) & IN-V2 Top-1 (\%)\\
\midrule
DualFormer-S18 (Ours) &\checkmark& \checkmark & 23 & 3.2 & \textbf{83.0} & \textbf{31.5} & \textbf{47.2} & 34.5 & \textbf{72.5} \\
DualFormer-S18 &\checkmark&  & 23 & 3.2 & \textbf{83.0} & \textbf{31.5} & 46.6 & 33.9 & 72.2\\
DualFormer-S18 &&\checkmark  & 23 & 3.2 & 82.9 & 30.6 & 47.0 & \textbf{34.6} & \textbf{72.5}\\
DualFormer-S18 &&  & 23 & 3.2 & 82.8 & 29.9 & 46.3 & 33.3 & 72.2\\

\bottomrule
\end{tabular}
}
\vspace{-2mm}
\end{table}

\end{document}